\documentclass[runningheads]{llncs}
\usepackage[T1]{fontenc}
\usepackage{graphicx}
\usepackage{booktabs}
\usepackage[misc]{ifsym}

\usepackage{amsmath}
\usepackage[hidelinks]{hyperref}
\usepackage{xcolor}
\usepackage{pifont}
\usepackage{amsfonts}       % blackboard math symbols
\usepackage{nicefrac}       % compact symbols for 1/2, etc.
\begin{document}

\title{STRIKE: Additive Feature-Group-Aware Stacking Framework for Credit Default Prediction}

\titlerunning{A Feature-Group-Aware Stacking Framework}

\newcommand{\equalcontrib}{\textsuperscript{*}}

\author{
Swattik Maiti\equalcontrib\inst{1} \and
Ritik Pratap Singh\equalcontrib\inst{1} \and
Fardina Fathmiul Alam\inst{2}
}

\authorrunning{S. Maiti, R. P. Singh, and F. F. Alam}

\institute{Science Academy, University of Maryland, College Park MD, USA \email{\{swat0507,ritik294\}@umd.edu}
\and
Department of Computer Science, University of Maryland, College Park MD, USA \email{fardina@umd.edu}}
% ==========================================================

\maketitle              % typeset the header of the contribution

\begin{center}
\small \textsuperscript{*} These authors contributed equally.
\end{center}

\begin{abstract}
Credit risk default prediction remains a cornerstone of risk management in the financial industry. The task involves estimating the likelihood that a borrower will fail to meet debt obligations, an objective critical for lending decisions, portfolio optimization, and regulatory compliance. Traditional machine learning models such as logistic regression and tree-based ensembles are widely adopted for their interpretability and strong empirical performance. However, modern credit datasets are high-dimensional, heterogeneous, and noisy, increasing overfitting risk in monolithic models and reducing robustness under distributional shift. We introduce STRIKE (Stacking via Targeted Representations of Isolated Knowledge Extractors), a feature-group-aware stacking framework for structured tabular credit risk data. Rather than training a single monolithic model on the complete dataset, STRIKE partitions the feature space into semantically coherent groups and trains independent learners within each group. This decomposition is motivated by an additive perspective on risk modeling, where distinct feature sources contribute complementary evidence that can be combined through a structured aggregation. The resulting group-specific predictions are integrated through a meta-learner that aggregates signals while maintaining robustness and modularity. We evaluate STRIKE on three real-world datasets spanning corporate bankruptcy and consumer lending scenarios. Across all settings, STRIKE consistently outperforms strong tree-based baselines and conventional stacking approaches in terms of AUC-ROC. Ablation studies confirm that performance gains stem from meaningful feature decomposition rather than increased model complexity. Our findings demonstrate that STRIKE is a stable, scalable, and interpretable framework for credit risk default prediction tasks.

\end{abstract}

\section{Introduction}
\label{sec:Introduction}

Accurately predicting whether a borrower will repay a loan is critical for responsible lending, portfolio management, and regulatory compliance. Prior to the 2007--08 financial crisis, many loans were issued without adequate risk assessment, and existing models failed to capture the true default risk of mortgage-backed securities. These shortcomings contributed to widespread financial losses and a collapse in market confidence \cite{fcic2011report}. In response, regulatory bodies introduced the Basel framework, a set of international banking regulations designed to strengthen risk management and financial stability \cite{basel2006framework,basel2011reforms}. Subsequently, the rise of machine learning introduced widely used models such as logistic regression, decision trees, and gradient-boosted ensembles into credit scoring pipelines. While these methods achieve strong predictive performance, they typically treat the feature space as a single monolithic entity. Modern credit datasets, however, are inherently heterogeneous: borrower information originates from multiple sources such as demographics, bureau records, delinquency histories, and loan vintage signals, each reflecting different aspects of default risk. Training a single model on the full feature space can mix these heterogeneous signals, allowing noise and redundant variables to obscure localized predictive patterns. This is particularly problematic in high-dimensional credit datasets,
where spurious interactions and feature interference can degrade robustness and interpretability. Foundational statistical and ensemble models \cite{altman1968financial,ohlson1980financial,breiman2001random,friedman2001greedy} often struggle with these modern complexities. Furthermore, recent comprehensive reviews \cite{alaka2018systematic,kirkos2015recent} show that such shortcomings can lead to inaccurate risk assessments, causing unfair loan denials for creditworthy individuals and risky approvals for potential defaulters. 

% Prior work \cite{alaka2018systematic,altman1968financial,breiman2001random,friedman2001greedy,kirkos2015recent,ohlson1980financial} shows that such shortcomings can lead to inaccurate risk assessments, causing unfair loan denials for creditworthy individuals and risky approvals for potential defaulters.

Recent research has explored convolutional neural networks (CNNs) \cite{qian2023soft} and ensemble methods such as stacking \cite{zhang2021ensemble} to enhance predictive performance. CNN-based credit models have demonstrated competitive results by automatically learning nonlinear feature interactions and hierarchical feature representations from borrower data. Stacking, in particular, combines multiple models by using the predictions of base learners as inputs for a meta-learner. Although effective, traditional stacking and CNN-based approaches still treat the input feature space as monolithic, without explicitly distinguishing between semantically different feature groups. This can dilute informative signals through feature interference and also limits interpretability, which is particularly important in credit risk modeling where transparent decision rationale is required. The predictions from these group-specific models are subsequently combined using a meta-learner, forming a structured ensemble that preserves the modular contributions of different feature sources.

% To address this limitation, we propose STRIKE (Stacking via Targeted Representations of Isolated Knowledge Extractors), a refined stacking approach that leverages semantic segregation of features to enhance the learning pipeline. STRIKE first isolates the input features into coherent conceptual groups, such as, but not limited to, demographics, repayment history, and transaction behavior. Each group is modeled independently using monolithic base learners, capturing local patterns with minimal interference \cite{wolpert1992stacked,ali2019group,bai2016hierarchical}. The predictions from these group-specific models are subsequently combined using a meta-learner, forming a structured ensemble that preserves the modular contributions of different feature sources. This
% design improves robustness in high-dimensional and imbalanced credit datasets while maintaining clearer attribution of predictive signals, an important property for interpretability and regulatory transparency
% in financial decision systems.

To address this limitation, we propose STRIKE (Stacking via Targeted Representations of Isolated Knowledge Extractors), a refined stacking approach that leverages semantic segregation of features to enhance the learning pipeline. STRIKE first isolates the input features into coherent conceptual groups, such as, but not limited to, demographics, repayment history, and transaction behavior. Each group is modeled independently using monolithic base learners, capturing local patterns with minimal interference \cite{ali2019group,bai2016hierarchical}. The predictions from these group-specific models are subsequently combined using a meta-learner\cite{wolpert1992stacked}, forming a structured ensemble that preserves the modular contributions of different feature sources. This design improves robustness in high-dimensional and imbalanced credit datasets while maintaining clearer attribution of predictive signals, an important property for interpretability and regulatory transparency in financial decision systems.

\textbf{Our key contributions are:}

% The remainder of this paper is organized as follows. Section \ref{sec:Related_Works} reviews related work in credit risk modeling, with an emphasis on orthodox stacking approaches. Section \ref{sec:Methods} details the proposed STRIKE methodology. Section \ref{sec:Experiments} presents experimental evaluations on $3$ real-world datasets, including comparisons with multiple prior methods and benchmark models. Finally, Section \ref{sec:Conclusion} concludes the paper with a summary of key findings and outlines potential directions for future research.

\begin{itemize}
    \item A framework, STRIKE, for robust credit default prediction that utilizes feature-group-aware stacking to handle high-dimensional, heterogeneous, and noisy tabular credit data.
    \item A structured modeling approach that partitions the feature space into semantically coherent groups, training independent base learners to capture localized patterns and minimize cross-group interference. This design is formally grounded in an additive modeling perspective.
    \item Demonstrated superior performance on three real-world credit risk datasets, significantly outperforming traditional machine learning baselines, recent deep learning models, and orthodox stacking methods in terms of AUC-ROC, while maintaining model scalability.
\end{itemize}

The code for our framework and experiments with our datasets are available at \textbf{https://tinyurl.com/strikecodebase}

% The remainder of this paper is organized as follows: Section~\ref{sec:Related_Works} reviews related work in credit scoring and ensemble methods. Section~\ref{sec:additive_perspective} establishes the mathematical foundation and additive modeling perspective. Section~\ref{sec:strike_framework} describes the STRIKE framework in detail. Section~\ref{sec:Experiments} presents our experimental setup and evaluates STRIKE against benchmark models. Finally, Section~\ref{sec:empirical_ablation} provides an empirical analysis and ablation studies to validate our architectural choices.

\section{Related Works}
\label{sec:Related_Works}

\textbf{Classical and Ensemble-Based Methods in Credit Scoring.} Credit risk prediction has long relied on classical statistical models such as logistic regression \cite{altman1968financial,ohlson1980financial} and discriminant analysis. With the rise of machine learning, ensemble techniques like random forests \cite{breiman2001random} and gradient boosting \cite{friedman2001greedy} became popular due to their ability to handle non-linearity and heterogeneous features. However, these models often struggle in high-dimensional credit datasets due to sensitivity to noise, limited handling of inter-feature dependencies, and challenges posed by class imbalance \cite{kirkos2015recent,alaka2018systematic}.

\textbf{Robustness to Outliers and Feature Redundancy.}
Outlier noise and redundant features further complicate risk classification. Approaches such as Local Outlier Factor (LOF) \cite{breunig2000lof}, isolation forests \cite{liu2008isolation}, and ensemble-based adaptations \cite{wei2019adaptive,zhang2021ensemble} have been proposed to reduce their impact. On the feature processing side, tree-based transformation techniques \cite{he2014statistical} and statistical selection methods like chi-square filtering help reduce redundancy, improving downstream classifier stability.

\textbf{Stacking for Risk Prediction.}
Stacking \cite{wolpert1992stacked} has emerged as a powerful ensemble strategy for credit scoring, particularly when incorporating model diversity across stages. Multi-stage variants \cite{zhang2021ensemble} aim to combine predictive strength and robustness through meta-learning, yet typically overlook the structural composition of features. This assumption of homogeneity often causes models to miss latent patterns that exist within specialized variable types.

\textbf{Towards Structure-Aware Modeling.}
Recent work emphasizes the benefits of incorporating domain-specific structure into learning pipelines. For example, separating variables by source or type—demographic, behavioral, transactional—can enhance both performance and interpretability \cite{ali2019group,bai2016hierarchical}. Although such feature grouping is intuitive in domains like finance, its integration within ensemble learning pipelines remains limited and largely underexplored.

In this context, our STRIKE methodology builds on existing ensemble and stacking literature \cite{zhang2021ensemble,stackingopt} by systematically integrating feature group partitioning into the model architecture. Rather than treating all features as equally informative, STRIKE isolates and models semantically coherent feature subspaces independently before fusing their predictions via a meta-learner. This structured approach directly addresses the noise sensitivity and feature interference seen in prior models, offering a scalable and modular solution for real-world credit risk prediction.

% =====================================================
\section{Methodology}
\label{sec:strike_framework}

\begin{figure}[t]
  \centering
  \includegraphics[width=\textwidth]{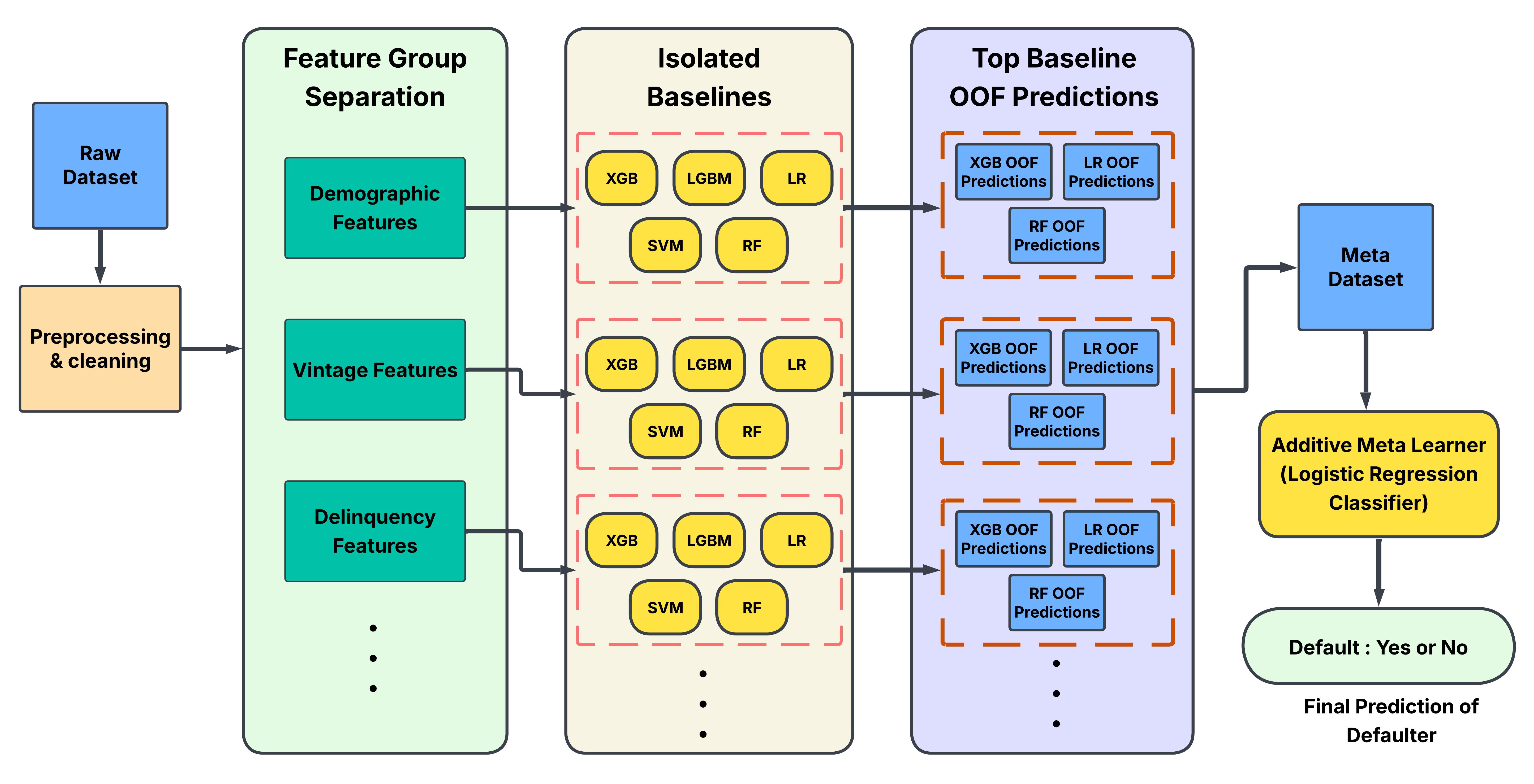}
  \caption{Overview of STRIKE. Features are partitioned into semantically coherent groups.
  Within each group, diverse base learners are trained using $K$-fold cross-validation to
  generate out-of-fold (OOF) predictions. Selected OOF predictions are concatenated to form
  a meta-dataset used to train a final meta-learner.}
  \label{fig:strike_overview}
\end{figure}

% \textbf{STRIKE} (Stacking via Targeted Representations of Isolated Knowledge Extractors)
% operationalizes the additive modeling perspective introduced in
% Section~\ref{sec:additive_perspective}.
% The framework separates feature groups during base model training
% and aggregates their predictions through a structured meta-learning stage.
Borrower information originates from heterogeneous sources that provide complementary evidence of default risk. To avoid the feature interference and overfitting inherent in monolithic models, STRIKE isolates these signals through a three-stage pipeline: semantic feature group separation, isolated baseline training, and additive meta-learner aggregation. An overview is shown in Fig. ~\ref{fig:strike_overview}.

\paragraph{\textbf{Feature Group Separation.}}
The full feature space is partitioned into semantically coherent groups.
In credit risk modeling, these groups may correspond to distinct data sources
such as demographics, bureau attributes, repayment history, transaction behavior,
or loan vintage. Grouping can be defined using domain knowledge or automated
heuristics by creating correlation-based or mutual-information-based feature clusters.
This decomposition enables each base learner to specialize within a more
homogeneous feature subspace, reducing interference from unrelated signals.

\paragraph{\textbf{Isolated baselines Training.}}
Within each feature group, STRIKE trains a diverse set of base models,
including tree-based ensembles such as XGBoost, LightGBM, and Random
Forest, as well as linear models such as Logistic Regression. Training
is performed using stratified $K$-fold cross-validation. At each fold,
models are trained on $K-1$ folds and generate predictions on the
held-out fold. Aggregating predictions across folds yields complete
out-of-fold (OOF) prediction vectors, which provide unbiased estimates
of model performance. Importantly, the cross-validation scores obtained at this stage offer a direct view of the predictive strength of each feature group, allowing
users to assess how different sources of borrower information contribute to overall risk prediction.

\begin{figure}[tb]
  \centering
  \includegraphics[width=0.8\textwidth]{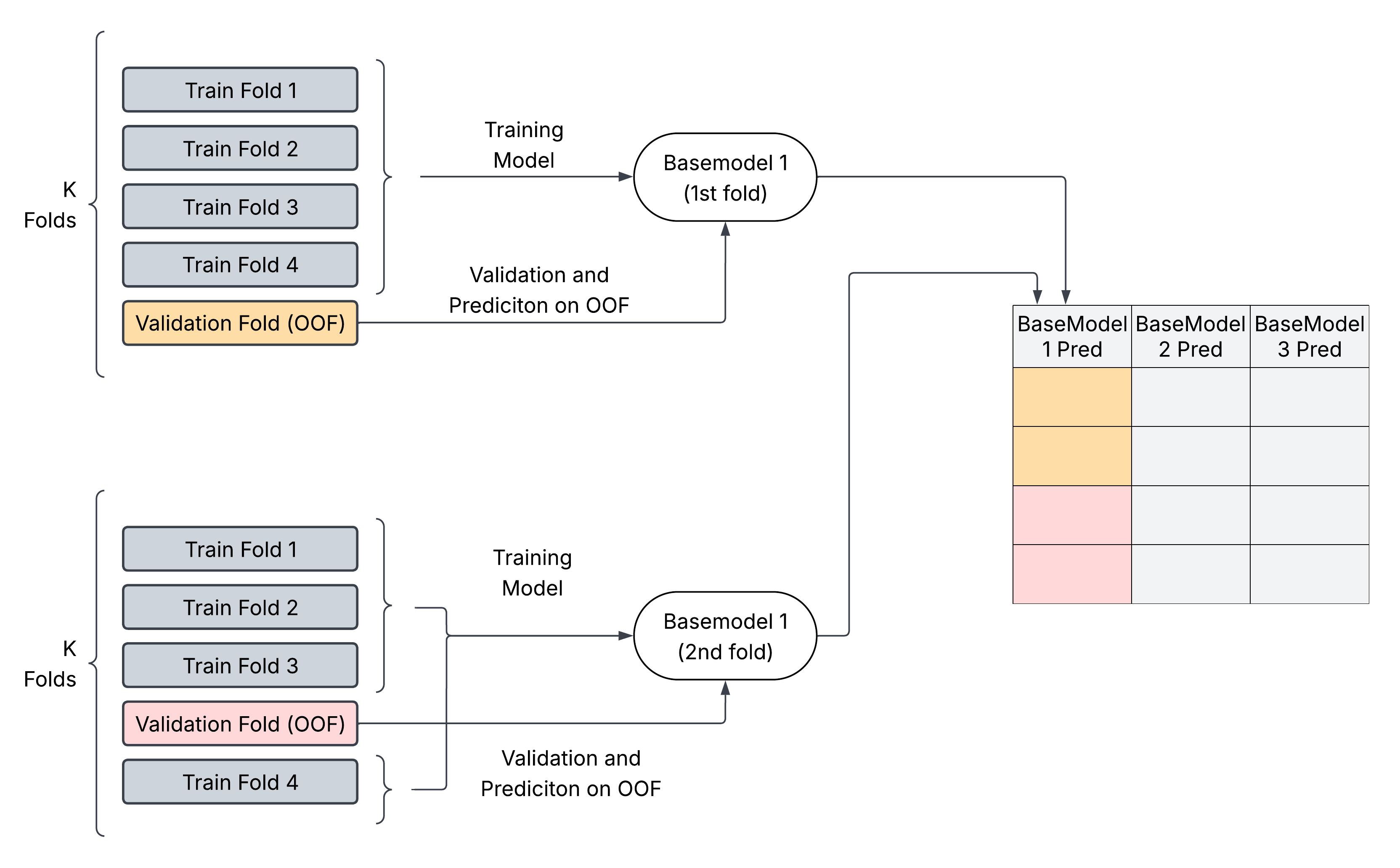}
  \caption{OOF prediction generation and meta-dataset creation in STRIKE. Stratified $K$-fold cross-validation produces leakage-free OOF predictions within each feature group. Validation-fold predictions are inserted into their indices to form complete OOF vectors, which are concatenated to build the meta-dataset for final training.}
  \label{fig:oof_diagram}
\end{figure}

\paragraph{\textbf{Meta-Dataset Construction.}}
For each feature group, the top-performing base models are selected
according to cross-validated AUC. Their OOF prediction vectors are
concatenated across groups to form a structured meta-dataset.
Because OOF predictions are generated from models that have not seen
the corresponding targets during training, the resulting meta-dataset
remains free from information leakage. An overview of this is illustrated in Fig.~\ref{fig:oof_diagram}.

\paragraph{\textbf{Meta-Learner Aggregation.}}
An additive meta learner like Logistic Regression Classifier is used as the default meta-learner to
combine groupwise predictions. By design, this additive aggregator learns to balance complementary signals across feature groups while inherently preventing over-complex, spurious cross-group interactions. Alternative meta-learners
can be substituted if desired; however, STRIKE’s primary gains arise
from structured feature decomposition rather than increased model
complexity.
% =====================================================
\section{Theoretical Foundation}
\label{sec:additive_perspective}

STRIKE is motivated by the observation that credit datasets often contain \emph{heterogeneous} feature sources
(e.g., demographics, bureau attributes, delinquency history, vintage/recency signals) that provide partially independent evidence about default risk. We formalize this intuition using an additive log-odds view of binary classification, which motivates
(i) training \emph{groupwise} predictors and (ii) combining them through an additive meta-learner.

% -----------------------------------------------------
\subsection{Additive log-odds decomposition}
\label{subsec:additive_logodds}

Let $Y \in \{0,1\}$ denote the default label and let the full feature vector be partitioned into
$G$ semantically coherent groups,
$X = \big(X^{(1)}, \ldots, X^{(G)}\big)$, where each $X^{(g)}$ corresponds to a feature block.
The Bayes-optimal classifier is the posterior $P(Y=1 \mid X=x)$.
It is convenient to work with its log-odds (logit),
\begin{equation}
f^\star(x)
\;:=\;
\log\frac{P(Y=1\mid X=x)}{P(Y=0\mid X=x)}.
\label{eq:bayes_logodds}
\end{equation}
By Bayes' rule, the posterior log-odds decomposes into a log-likelihood ratio plus a prior offset:
\begin{equation}
\log\frac{P(Y=1\mid X)}{P(Y=0\mid X)}
\;=\;
\log\frac{P(X\mid Y=1)}{P(X\mid Y=0)}
\;+\;
\log\frac{P(Y=1)}{P(Y=0)}.
\label{eq:logodds_decomp}
\end{equation}

\paragraph{\textbf{Conditional independence across groups.}}
A natural structural assumption is that feature groups are approximately conditionally independent
given the outcome:
\begin{equation}
\textbf{(A1)} \qquad
X^{(g)} \perp X^{(h)} \mid Y, \quad \forall\, g \neq h .
\label{eq:A1}
\end{equation}
Under \eqref{eq:A1}, the class-conditional likelihood factorizes,

$P(X\mid Y=y) = \prod_{g=1}^G P\!\left(X^{(g)}\mid Y=y\right)$,
and plugging this into \eqref{eq:logodds_decomp} yields an additive decomposition of the Bayes log-odds:
\begin{equation}
f^\star(x)
\;=\;
\sum_{g=1}^G
\underbrace{
\log\frac{P\!\left(x^{(g)}\mid Y=1\right)}{P\!\left(x^{(g)}\mid Y=0\right)}
}_{=:~f_g^\star(x^{(g)})}
\;+\; C,
\qquad
C := \log\frac{P(Y=1)}{P(Y=0)} .
\label{eq:additive_bayes}
\end{equation}
Equation \eqref{eq:additive_bayes} suggests modeling each feature group with a specialized predictor
and combining group evidence additively.

% -----------------------------------------------------
\subsection{Groupwise predictors and a logistic meta-learner}
\label{subsec:logistic_meta}

In practice, the groupwise Bayes components $f_g^\star$ are unknown.
STRIKE trains base learners within each group to estimate group-level default probabilities
$\hat{p}_g(x^{(g)}) \approx P(Y=1\mid X^{(g)}=x^{(g)})$.
We convert these to group logits
$\ell_g(x^{(g)}) := \log\!\big(\hat{p}_g(x^{(g)})/(1-\hat{p}_g(x^{(g)}))\big)$
and combine them using a logistic regression meta-model:
\begin{equation}
\text{logit}\,\hat{p}(x)
\;=\;
\beta_0 \;+\; \sum_{g=1}^G \beta_g\,\ell_g\!\left(x^{(g)}\right).
\label{eq:weighted_additive}
\end{equation}
This implements a \emph{data-driven weighted additive rule}: redundant or noisy groups can be
downweighted ($\beta_g \approx 0$), while uniquely informative groups receive larger weights.

% -----------------------------------------------------
\subsection{Residual interactions and inductive bias}
\label{subsec:residual_interactions}

Assumption \eqref{eq:A1} is an approximation: credit features can exhibit genuine cross-group
dependencies (e.g., repayment delinquencies interacting with recency/vintage effects).
When \eqref{eq:A1} is violated, the Bayes log-odds admits an additive component plus a residual
interaction term:
\begin{equation}
f^\star(x)
\;=\;
\sum_{g=1}^G f_g^\star(x^{(g)})
\;+\;
\delta(x)
\;+\; C,
\qquad
\delta(x)
\;:=\;
\log\frac{P(X\mid Y)}{\prod_{g=1}^G P\!\left(X^{(g)}\mid Y\right)} .
\label{eq:residual_delta}
\end{equation}
STRIKE treats additivity as a \emph{default inductive bias}: base learners specialize within groups
to reduce variance in high-dimensional and noisy settings, while the meta-learner provides a
low-capacity correction mechanism that can absorb a limited amount of cross-group dependence through
its learned weights.
In Section~\ref{sec:empirical_ablation}, we empirically analyze cross-group dependence and ablate the
impact of grouping choices and meta-learner capacity.

\subsection{Empirical Justification of Conditional Independence}
\label{subsec:dependence}

A central premise of our additive log-odds decomposition is that feature groups provide conditionally independent evidence about the target outcome. If cross-group dependence is pervasive, decomposing the feature space might discard important interactions. Conversely, if the dependence is limited (as our inductive bias assumes), groupwise modeling effectively isolates predictive signals while reducing noise and overfitting. To validate this structural assumption empirically, we measure the conditional dependence between feature groups using Conditional Mutual Information (CMI).

\begin{figure}[t]
\centering
\includegraphics[width=0.5\linewidth]{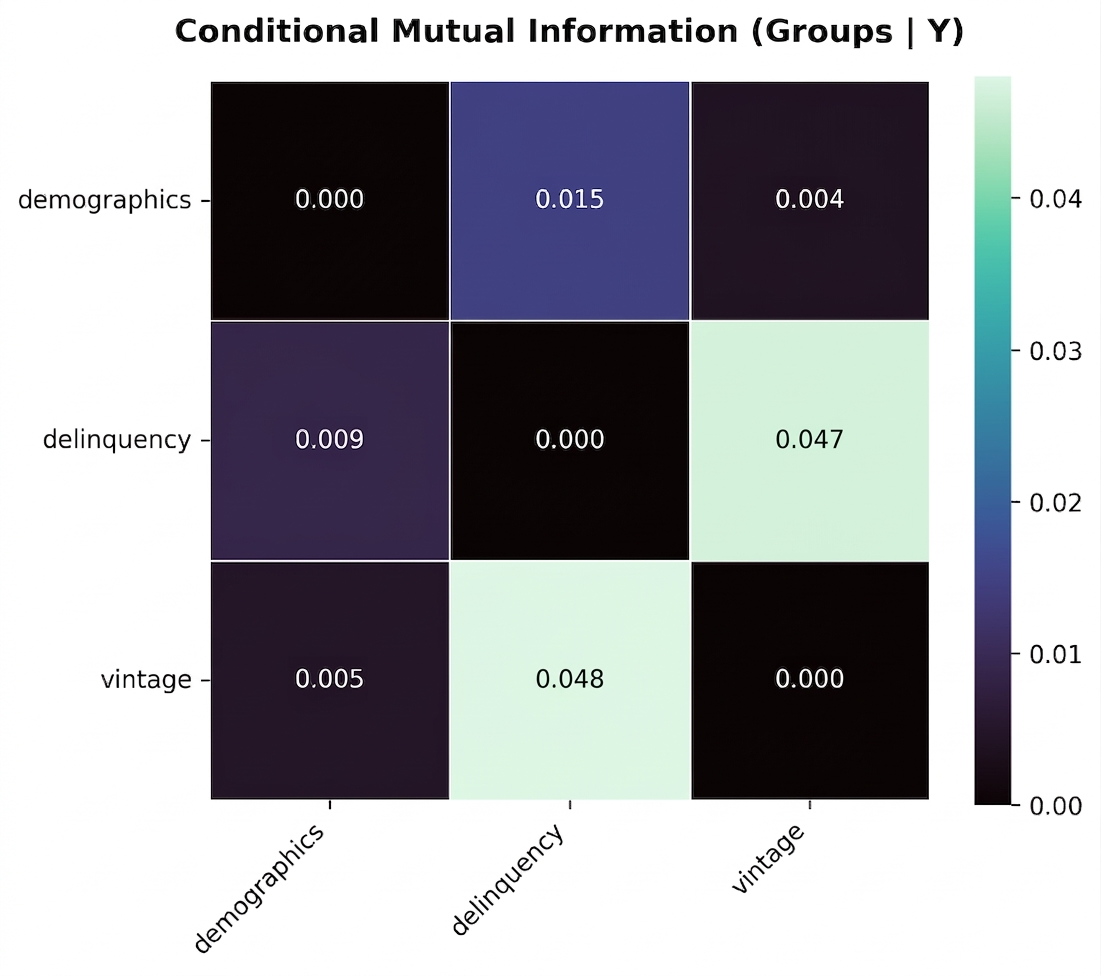}
\caption{Conditional Mutual Information $I(X^{(g)};X^{(h)}\mid Y)$ between feature groups evaluated on the HomeCredit dataset. The uniformly low off-diagonal values empirically support the approximate conditional independence assumption (Eq.~\ref{eq:A1}), which motivates the additive log-odds decomposition underlying the STRIKE framework.}
\label{fig:cmi_heatmap}
\end{figure}

\paragraph{\textbf{Conditional Mutual Information (CMI).}}For feature groups $X^{(g)}$ and $X^{(h)}$, CMI measures dependence after conditioning on $Y$:
$I\!\left(X^{(g)}; X^{(h)} \mid Y\right)$.
Low values indicate that, once the outcome is fixed, the two groups contain little to redundant information about default.
Figure~\ref{fig:cmi_heatmap} shows that off-diagonal CMI values are uniformly small,
with a mean of $0.022$ across group pairs from the HomeCredit Dataset.
This suggests that groupwise signals are largely non-redundant after conditioning on $Y$,
supporting STRIKE’s bias toward additive aggregation. This observation aligns with STRIKE’s design: base learners model group-specific signals independently, while the meta-stage aggregates their predictions to capture residual dependencies across groups.

\section{Experiments}
\label{sec:Experiments}

% To verify the effectiveness of the proposed STRIKE framework, we conduct comprehensive experiments on multiple real-world credit risk datasets of varying sizes and characteristics. Section~\ref{sec:Datasets-Preprocessing} presents the basic information of the utilized datasets along with the preprocessing methods. Section~\ref{sec:Compute Resources} explains the compute resources used in the experiments. Section~\ref{sec:Experimental-Setup} introduces the experimental settings in detail, along with data splitting and model training. Section~\ref{sec:Evaluation-Metrics} describes the evaluation metrics employed for assessing model performance. Finally, Section~\ref{sec:Benchmark-Comparison} provides a comparative analysis of STRIKE against standard benchmark algorithms. Through these experiments, we aim to demonstrate the robustness, scalability, and predictive superiority of STRIKE across diverse credit scoring scenarios.

We verify STRIKE’s effectiveness via experiments on $3$ real-world credit risk datasets.  
All experiments were run on a High Performance compute cluster (40 vCPUs, 8 GB RAM per vCPU, 1.5 TB local + 1.5 PB BeeGFS scratch) using Python 3.10 on SLURM (no GPU). 
Section~\ref{sec:Datasets-Preprocessing} details dataset characteristics and preprocessing. 
Section~\ref{sec:Experimental-Setup} outlines experimental settings including data splitting and model training, and Section~\ref{sec:Evaluation-Metrics} presents evaluation metrics.  
Finally, Section~\ref{sec:Benchmark-Comparison} provides a comparative analysis of STRIKE against standard benchmark algorithms.

\subsection{Datasets Description and Preprocessing}
\label{sec:Datasets-Preprocessing}

% We evaluate the STRIKE framework across the three real-world credit scoring datasets: \textbf{Polish} \cite{uci_polish_bankruptcy}, \textbf{LendingClub} \cite{lendingclub_dataset}, and \textbf{HomeCreditDefaultRisk} \cite{homecredit2018}. These datasets are widely used benchmarks in recent credit risk modeling research, including the SR1D-CNN study~\cite{qian2023soft} enabling direct comparisons between STRIKE and recent deep learning approaches while offering diverse challenges for credit risk modeling. The Polish dataset is small and highly imbalanced (3.86\% bankruptcies), ideal for testing models under limited data. LendingClub provides a moderate imbalance (21.3\% defaults) across 225,611 consumer loan records, enabling real-world peer-to-peer lending analysis. HomeCreditDefaultRisk presents the most complex scenario, with 307,511 samples, 597 sparse and noisy features, and an 8.07\% default rate, serving as a robust testbed for scalability and handling data imperfections. For more details, see Appendix~\ref{app:dataset}. Notably, the selected datasets span a diverse range of credit risk scenarios, including peer-to-peer lending, corporate bankruptcy prediction, and consumer credit default, allowing for a comprehensive assessment of the generalizability of STRIKE in varied financial contexts. 

In this study, we evaluate the STRIKE framework across three real-world credit scoring datasets: \textbf{Polish}\cite{uci_polish_bankruptcy}, \textbf{LendingClub}\cite{lendingclub_dataset}, and \textbf{HomeCreditDefaultRisk}\cite{homecredit2018}. These datasets are widely used benchmarks in recent credit risk modeling research, including the SR1D-CNN study~\cite{qian2023soft}, enabling direct comparisons between STRIKE and recent deep learning approaches. Notably, the selected datasets span a diverse range of credit risk scenarios—including peer-to-peer lending, corporate bankruptcy prediction, and consumer credit default—allowing for a comprehensive assessment of STRIKE’s generalizability across varied financial contexts. To ensure consistency and isolate STRIKE’s modeling capabilities, all datasets underwent identical preprocessing: categorical features were one-hot encoded, missing values were set to $-999$, and features were scaled to the $[0,1]$ range using min-max normalization. Table~\ref{dataset-description-table} summarizes key dataset characteristics.

\begin{table}[htbp]
  \caption{Summary of datasets used for evaluating STRIKE. The table reports the sample size, class distribution (number of positive samples representing defaults and negative samples representing non-defaults), and the number of raw features before any preprocessing.}
  \label{dataset-description-table}
  \centering
  \begin{tabular}{lcccc}
    \toprule
    \textbf{Dataset} & \textbf{Sample Size } & \textbf{Positive Samples } & \textbf{Negative Samples } & \textbf{Features} \\
    \midrule
    Polish                  &   7,027   &    271    &   6,756    & 65  \\
    LendingClub             & 225,611   & 48,015    & 177,596    & 150 \\
    HomeCredit   & 307,511   & 24,825    & 282,686    & 597 \\
    \bottomrule
  \end{tabular}
\end{table}

\textbf{Polish:} \cite{uci_polish_bankruptcy}
This dataset focuses on forecasting the bankruptcy of Polish companies using financial ratios derived from balance sheets and income statements. Comprising 7,027 records with only 3.86\% representing bankrupt companies, the dataset’s limited size and pronounced class imbalance present ideal conditions for evaluating model robustness under constrained data scenarios.

\textbf{LendingClub Dataset:} \cite{lendingclub_dataset}
Unlike the Polish dataset, which focuses on company-level bankruptcy, the LendingClub dataset centers on individual-level peer-to-peer lending risk. Originating from a U.S.-based lending platform, it includes 225,611 loan records with 150 features detailing borrower information, loan characteristics, and financial history. It exhibits a moderate class imbalance, with defaults constituting about 21.3\% of the data. The dataset's real-world lending features and moderate imbalance provide a practical environment for assessing model performance in consumer credit risk prediction.

\textbf{HomeCreditDefaultRisk:} \cite{homecredit2018} Released by the Home Credit Group, this high-dimensional dataset contains 307,511 samples and 597 features which exhibits significant sparsity and noise. With a default rate of approximately 8.07\%, it presents a challenging testbed for evaluating model performance under extreme class imbalance, missing data, and scalability constraints.

To ensure consistency and isolate STRIKE’s modeling capabilities, all datasets underwent identical preprocessing: categorical features were one-hot encoded, missing values were set to $-999$, and features were scaled to the $[0,1]$ range using min-max normalization. 

Features were partitioned by domain into specific groups: the Polish dataset into \textbf{Profitability}, \textbf{Leverage}, \textbf{Liquidity}, \textbf{Efficiency}, and \textbf{Growth}; the Lending Club dataset into \textbf{Loan Terms}, \textbf{Credit Profile}, \textbf{Utilization}, and \textbf{Categorical Flags}; and the Home Credit dataset into \textbf{Demographics}, \textbf{Vintage}, and \textbf{Delinquency}.

\subsection{Experimental Setup and Model Training}
\label{sec:Experimental-Setup}

All experiments were conducted using Python, with the primary codebase developed in Visual Studio Code. Model training and evaluation were performed on high-performance cloud cluster compute environments to ensure efficient experimentation across large datasets. For each dataset, an initial split was performed to separate the data into training and testing subsets using a 70\%--30\% stratified division. The training set was then subjected to 5-fold stratified cross-validation during the base model training phase to generate out-of-fold (OOF) predictions while maintaining a fair validation structure.

For baseline model training within each feature group, a consistent approach was adopted across datasets: five diverse tree-based models were selected from the pool of candidate algorithms, including XGBoost, LightGBM, CatBoost, AdaBoost, Random Forest, ExtraTrees, and Gradient Boosting Decision Trees (GBDT). The specific five models chosen varied slightly across datasets but generally emphasized model diversity within the ensemble. After training, the top three performing base models for each feature group were selected based on their cross-validated AUC scores. These selected models' OOF predictions were used to construct the meta-dataset.

For final prediction, a logistic regression meta-learner was trained on the meta-dataset constructed from the training folds. During testing, the same sequence was followed: the pre-trained base models were loaded from stored pickle files to generate the testing meta-dataset, which was then passed through the pre-trained logistic regression meta-learner to produce the final output predictions. This experimental pipeline was applied consistently across all three datasets to ensure a fair and comparable evaluation of STRIKE's performance.

%end of experimental setup and model training

\subsection{Model Evaluation Metrics}
\label{sec:Evaluation-Metrics}

In credit risk classification tasks, multiple evaluation metrics can be used depending on the specific business objectives and class imbalance considerations. While accuracy is often reported, it may be misleading in highly imbalanced datasets. To address this, we  compute the F1-score, which balances precision and recall. However, for the purpose of general model comparison and consistency with recent literature in credit scoring, we adopt the Area Under the Receiver Operating Characteristic Curve (AUC-ROC)~\cite{fawcett2006introduction} as our primary evaluation metric. AUC-ROC provides a threshold-independent measure of a model’s ability to distinguish between positive and negative classes, which is critical for assessing model robustness in imbalanced settings.

The Area Under the ROC Curve (AUC) can be formally expressed as:

\[
\text{AUC} = \int_{0}^{1} \text{TPR}(\text{FPR}) \, d(\text{FPR})
\]

Here, TPR (Sensitivity) and FPR (1 - Specificity) are functions of the decision threshold, and the AUC represents the probability that a randomly chosen positive instance is ranked higher than a randomly chosen negative instance.

\subsection{Comparison with Benchmark Models}
\label{sec:Benchmark-Comparison}

\paragraph{\textbf{Comparison with traditional and deep learning models.}  }
Table~\ref{tab:model-comparison-table-selected} benchmarks STRIKE against widely-used models in the credit scoring domain, including Logistic Regression, Random Forest, AdaBoost, GBDT, and deep learning approaches such as DeepFM, DCN-V2, and SR1D-CNN~\cite{qian2023soft}. While tree-based ensembles and CNN variants like SR1D-CNN demonstrate solid performance, STRIKE consistently delivers superior AUC-ROC scores across all datasets, notably outperforming SR1D-CNN by over 15 percentage points on the Polish dataset.

Importantly, for this comparison, all models inside the STRIKE architecture were trained using default parameters without any hyperparameter tuning. This design choice was intentional: it isolates the architectural advantage of STRIKE and showcases its raw predictive power, independent of fine-tuning. STRIKE addresses a core limitation of SR1D-CNN: its reliance on convolutional layers that assume spatial coherence among input features. However, this assumption stems from inductive biases like spatial local correlation and weight sharing, which are effective in domains like image and text processing but do not hold for tabular credit scoring data where adjacent features lack such spatial relationships~\cite{qian2023soft,grinsztajn2022tree}. STRIKE avoids this pitfall by explicitly grouping features based on domain semantics (e.g., demographic, delinquency, vintage, amount financed), and learning localized representations before combining them in a controlled, modular fashion. This enables STRIKE to generalize better in scenarios marked by heterogeneous feature types, missing values, and non-linear borrower behaviors—key challenges where conventional deep architectures often struggle. STRIKE’s modular inductive bias thus provides a more faithful and robust approach to modeling creditworthiness across varied credit risk datasets.

\begin{table}[h]
\centering
  \caption{\textbf{Performance comparison of STRIKE with other machine learning methods on three datasets.}
  The table reports the cross-validated AUC-ROC (mean $\pm$ standard deviation) achieved by each algorithm on the Polish, LendingClub, and HomeCreditDefaultRisk datasets. \textit{Note: Bold font highlights the best AUC-ROC values in each column}}
  \label{tab:model-comparison-table-selected}
  \begin{tabular}{lccc}
    \toprule
    \textbf{Model} & \textbf{Polish} & \textbf{LendingClub} & \textbf{HomeCredit} \\
    
    \midrule
    \multicolumn{4}{c}{\textbf{Traditional Baselines}} \\
    \midrule
    LR & 0.5300$\pm$0.0059 & 0.7144$\pm$0.0001 & 0.7392$\pm$0.0002 \\
    DT & 0.7532$\pm$0.0128 & 0.6257$\pm$0.0012 & 0.6073$\pm$0.0011 \\
    
    \midrule
    \multicolumn{4}{c}{\textbf{Tree-Based Ensembles}} \\
    \midrule
    AdaBoost & 0.8707$\pm$0.0126 & 0.7199$\pm$0.0002 & 0.7486$\pm$0.0003 \\
    RF & 0.8704$\pm$0.0058 & 0.7000$\pm$0.0011 & 0.7091$\pm$0.0026 \\
    GBDT & 0.9352$\pm$0.0038 & 0.7217$\pm$0.0001 & 0.7485$\pm$0.0003 \\
    XGBoost & 0.9508$\pm$0.0038 & 0.7214$\pm$0.0001 & 0.7475$\pm$0.0002 \\
    LightGBM & 0.9518$\pm$0.0038 & 0.7217$\pm$0.0002 & 0.7489$\pm$0.0003 \\
    CatBoost & 0.9443$\pm$0.0025 & 0.7190$\pm$0.0002 & 0.7453$\pm$0.0003 \\
    
    \midrule
    \multicolumn{4}{c}{\textbf{Deep Learning / Neural Networks}} \\
    \midrule
    MLP & 0.5340$\pm$0.0143 & 0.7008$\pm$0.0019 & 0.6981$\pm$0.0029 \\
    DeepFM & 0.5659$\pm$0.0056 & 0.7177$\pm$0.0006 & 0.7431$\pm$0.0019 \\
    DCN-V2 & 0.5656$\pm$0.0181 & 0.7180$\pm$0.0007 & 0.7429$\pm$0.0010 \\
    TabNet & 0.5357$\pm$0.0064 & 0.7217$\pm$0.0029 & 0.7437$\pm$0.0020 \\
    1D CNN & 0.5235$\pm$0.0077 & 0.7209$\pm$0.0007 & 0.7410$\pm$0.0022 \\
    SR-1D-CNN & 0.8149$\pm$0.0174 & 0.7271$\pm$0.0004 & 0.7517$\pm$0.0007 \\
    
    \midrule
    \multicolumn{4}{c}{\textbf{Proposed Framework}} \\
    \midrule
    \textbf{STRIKE} & \textbf{0.9683$\pm$0.0020} & \textbf{0.7503$\pm$0.0003} & \textbf{0.7661$\pm$0.0008} \\
    \bottomrule
  \end{tabular}
\end{table}

\paragraph{\textbf{Comparison with orthodox stacking.}}
We also compare the performance of STRIKE with a traditional stacking ensemble proposed by Zhang et al.~\cite{zhang2021ensemble}, which trains all base learners on the full feature space without feature group isolation. To ensure a fair comparison, we use the same base classifiers for both methods and conduct the evaluation under an identical experimental setup on the Polish dataset. As expected, the standalone performance of base learners across both setups appears comparable in terms of accuracy, balanced accuracy, and log loss; underscoring that the base models themselves are not inherently superior in either method.

\begin{figure}[tb]
  \centering
  \begin{minipage}[t]{0.45\textwidth}
    \centering
    \includegraphics[width=\textwidth]{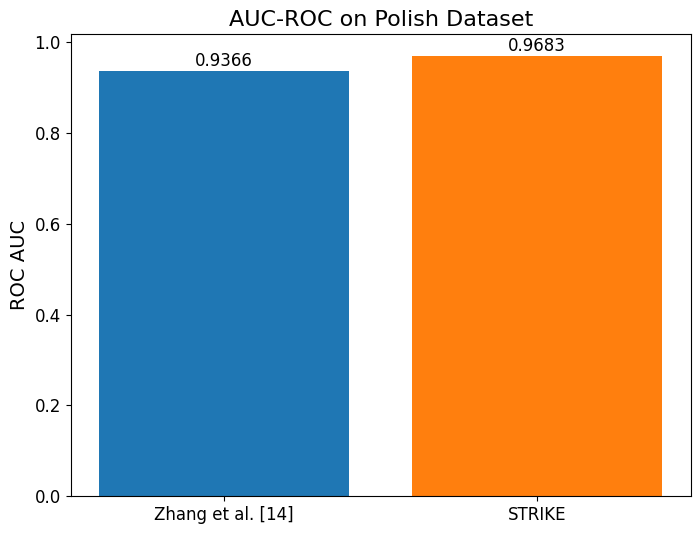}
    \caption{
    %\textbf{AUC-ROC comparison between STRIKE and Orthodox Stacking.}
    AUC-ROC performance: STRIKE vs.\ Orthodox Stacking.
    \textit{
    % This bar chart illustrates the difference in AUC-ROC performance on the Polish dataset, where STRIKE demonstrates a clear advantage over the ensemble model proposed by~\cite{zhang2021ensemble}.
% This bar chart shows STRIKE outperforms the orthodox stacking ensemble of Zhang et al.~\cite{zhang2021ensemble} on the Polish dataset, achieving a substantially higher AUC-ROC.
    }
    }
    \label{fig:auc_roc_ref}
  \end{minipage}
  \hfill
  \begin{minipage}[t]{0.45\textwidth}
    \centering
    \includegraphics[width=\textwidth]{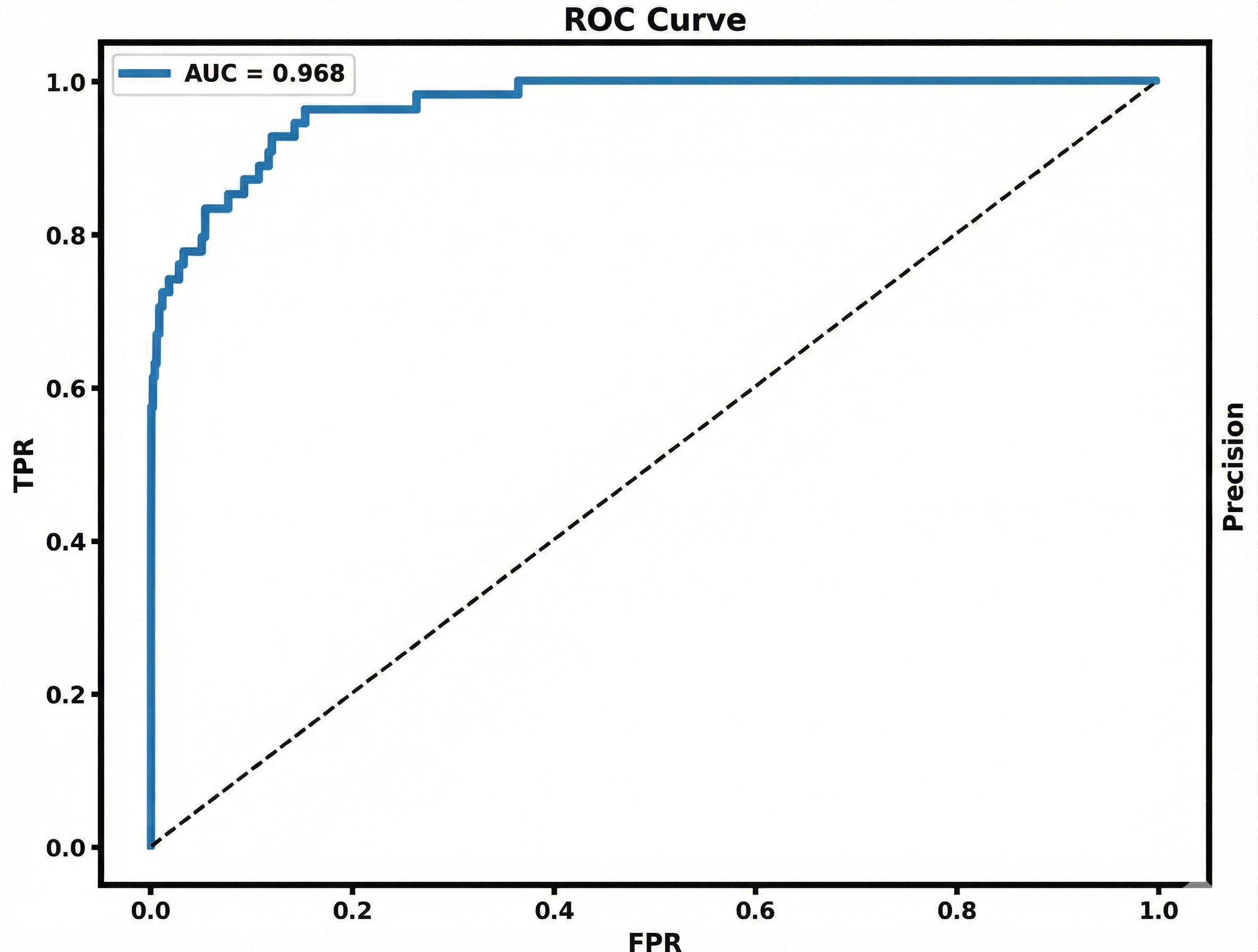}
    \caption{
    ROC Curve for STRIKE on the Polish dataset.
    % \textit{The ROC curve shows STRIKE’s strong classification ability, with an AUC of 0.968.}
    }
    \label{fig:auc_roc_ours}
  \end{minipage}
\end{figure}

However, the key advantage of STRIKE becomes evident when these base predictions are aggregated and passed to the meta learner. As shown in Figure~\ref{fig:auc_roc_ref}, STRIKE demonstrates a significant improvement of approximately \textbf{3.4\%} in final AUC score over the orthodox stacking model. This performance leap can be attributed to STRIKE’s core design principle: by isolating semantically distinct feature groups and training dedicated models per group, the architecture avoids the noise and redundancy that often plague full-feature approaches. The meta learner thus learns from cleaner, more disentangled prediction signals, enabling stronger generalization in credit default classification.

% =====================================================
\section{Empirical Analysis and Ablations}
\label{sec:empirical_ablation}

% The results in Section~\ref{sec:Experiments} show that STRIKE improves predictive performance
% across diverse credit risk settings. In this section, we investigate \emph{why} STRIKE works,
% \emph{when} it is most beneficial, and \emph{which} design choices matter.
% Our analysis is guided by the additive modeling perspective in
% Section~\ref{sec:additive_perspective}: STRIKE is expected to be most effective when
% (i) feature groups provide largely complementary evidence about default, and
% (ii) cross-group dependence exists but is \emph{sparse} rather than pervasive.
% We therefore (a) quantify cross-group dependence and interaction strength,
% and (b) perform controlled ablations that isolate the effect of grouping and meta-learning.

The results in Section ~\ref{sec:Experiments} show that STRIKE improves predictive performance across diverse credit risk settings. Having empirically validated the foundational assumption of cross-group independence in Section~\ref{sec:additive_perspective}, we now investigate the specific design choices that drive this performance. We perform controlled ablations to isolate the effect of feature grouping strategies and meta-learning capacity, ensuring the framework's gains stem from structured feature decomposition rather than arbitrary model configurations.

\subsection{Sensitivity to Group Definitions}
\label{subsec:ablation_groupdefs}

A natural question is whether STRIKE’s gains depend critically on a particular
manual grouping. To test robustness, we repeat STRIKE under alternative
grouping strategies that do not rely on domain knowledge, including
correlation-based clustering and mutual-information-based clustering.
Across these structured grouping strategies, results remain stable
(within $\Delta$AUC $< 0.01$), suggesting that STRIKE is not overly
sensitive to reasonable group definitions and that its benefits arise
from explicit feature decomposition rather than any single handcrafted
partition.

To further stress-test the method, we construct a \emph{random grouping}
baseline in which all features are pooled and assigned to groups using a
randomized round-robin procedure. Features are first shuffled using a
fixed seed and then distributed sequentially across groups so that each
group contains approximately the same number of features, producing
mixed groups that intentionally ignore semantic structure.
These experiments are conducted on the HomeCredit dataset using a
50k-row subset of the training data to allow repeated ablation trials.

\begin{table}[ht]
\centering
\caption{Ablation study evaluating sensitivity to feature grouping strategies on a 50k-row subset of the HomeCredit dataset. It compares the cross-validated AUC of structured groupings (manual, mutual information, correlation-based) against randomized mixed feature groupings. \textit{Note: Bold font highlights the best-performing value in META AUC Column and Highest change in delta vs Manual Column}}
\label{tab:grouping_sensitivity}
\begin{tabular}{lcc}
\toprule
\textbf{Grouping Strategy} & \textbf{Meta AUC (CV)} & \textbf{$\Delta$ vs. Manual} \\
\midrule
\multicolumn{3}{c}{\textit{Logical / Structured Grouping}} \\
\midrule
Manual (Domain Knowledge) & \textbf{0.7473} & -- \\
Mutual Information (MI) & 0.7467 & -0.0006 \\
Correlation-Based (CORR) & 0.7407 & -0.0066 \\
\midrule
\multicolumn{3}{c}{\textit{Random Mixed Feature Grouping}} \\
\midrule
Random Grouping (Seed 0) & 0.7388 & -0.0085 \\
Random Grouping (Seed 1) & 0.7380 & -0.0093 \\
Random Grouping (Seed 2) & 0.7392 & -0.0081 \\
Random Grouping (Seed 3) & 0.7367 & \textbf{-0.0106} \\
Random Grouping (Seed 4) & 0.7389 & -0.0084 \\
\midrule
Random Grouping (Mean) & 0.7383 & -0.0090 \\
\bottomrule
\end{tabular}
\end{table}

Table~\ref{tab:grouping_sensitivity} summarizes the results. Structured
groupings (manual or MI-based) yield very similar performance, while
correlation-based grouping produces only a modest drop. In contrast,
randomly mixing features consistently degrades performance, with the
mean AUC falling to 0.7383. Notably, this performance is even lower than
that of monolithic models trained on the same 50k sample, indicating that
arbitrary feature partitions disrupt useful signal structure rather than
improving specialization. These results suggest that while STRIKE does
not require a specific handcrafted grouping, preserving some underlying
structure in the feature decomposition is important for stable gains.

% -----------------------------------------------------

\subsection{Meta-Learner Choice}
\label{subsec:ablation_meta_capacity}

Having examined the role of feature grouping, we next study the choice of
meta-learner used to combine groupwise predictions. STRIKE uses a logistic regression combiner by default.
This choice aligns with the additive bias discussed in
Section~\ref{sec:additive_perspective} and reduces the risk of
overfitting on the meta-dataset.
To evaluate whether performance gains arise primarily from increased
meta-model capacity, we compare logistic regression with more flexible
additive meta-learners trained on the same meta-features.

Table~\ref{tab:meta_learner_ablation} summarizes the results on the
HomeCredit dataset. Logistic regression already provides strong
performance (0.7661 AUC), outperforming all monolithic baselines
reported earlier. Replacing it with more flexible additive meta-learners
further improves performance, with GAM achieving the best result
(0.7714 AUC) and EBM yielding a similar score (0.7713). These improvements indicate that modestly richer additive meta-models
can better capture residual nonlinearities in the combination of
groupwise predictions. Importantly, the gains arise without requiring
dense interaction modeling, suggesting that most predictive signal is
still captured through additive aggregation of groupwise models,
consistent with the inductive bias underlying STRIKE.

\begin{table}[ht]
\centering
\caption{Ablation study on the effect of meta-learner capacity using the HomeCredit dataset. It compares the performance of STRIKE's default logistic regression meta-learner against more flexible additive models (GAM and EBM). \textit{Note : Bold font highlights the best-performing value in AUC (CV) Column}}
\label{tab:meta_learner_ablation}
\begin{tabular}{lccc}
\toprule
\textbf{Meta-Learner} & \textbf{Additive} & \textbf{Interactions} & \textbf{AUC (CV)} \\
\midrule
Logistic Regression & \checkmark & -- & 0.7661 \\
GAM (Spline-based) & \checkmark & -- & \textbf{0.7714} \\
EBM (Sparse Interactions) & \checkmark & \checkmark & 0.7713 \\
\bottomrule
\end{tabular}
\end{table}

% -----------------------------------------------------
% \subsection{When STRIKE Helps Most}
% \label{subsec:when_strike_works}

% The above diagnostics and ablations provide a practical characterization of STRIKE’s operating regime.
% STRIKE is most beneficial when: (i) features can be partitioned into semantically meaningful sources,
% (ii) groupwise signals are complementary (low redundancy after conditioning on $Y$),
% and (iii) cross-group interactions exist but are localized rather than pervasive.
% In contrast, when feature groups are strongly entangled or engineered to be highly overlapping,
% monolithic full-feature models may be competitive, and STRIKE’s advantage can diminish.

% Overall, STRIKE should be viewed as a structured ensemble method that
% \emph{defaults to additivity} and \emph{regulates} interactions through a low-dimensional meta stage,
% offering improved stability and generalization in heterogeneous credit risk datasets.

\section{Conclusion}
\label{sec:final_conclusion}

This paper introduced STRIKE, a feature-group-aware stacking framework for
credit default prediction designed to address the heterogeneous structure of
modern credit datasets. Rather than training a single monolithic model over the
entire feature space, STRIKE partitions features into semantically coherent
groups, trains specialized base learners within each group, and aggregates
their predictions through a meta-learner. This design is motivated by an
additive log-odds perspective, where different feature sources provide
complementary evidence about default risk. Under approximate conditional
independence across groups, such a decomposition allows predictive signals to
be learned more robustly while reducing cross-feature interference.

% Empirical results across three real-world credit risk datasets demonstrate that
% STRIKE consistently improves AUC performance relative to traditional machine
% learning models, deep learning approaches, and orthodox stacking ensembles.
% Ablation studies further show that the gains stem from structured feature
% decomposition rather than increased model complexity. In particular, our
% analysis indicates that STRIKE performs best when feature groups contain
% complementary but weakly redundant information, a regime supported by the low
% cross-group conditional mutual information observed in the HomeCredit dataset.
% The framework also provides improved transparency compared to monolithic
% models, since groupwise base learners and their cross-validated performance
% offer a direct view of how different feature sources contribute to risk
% prediction. This modular structure makes the model easier to analyze and adapt
% in regulated financial environments where interpretability is important.
Empirical results across three real-world credit risk datasets demonstrate that
STRIKE consistently improves AUC performance relative to traditional machine
learning models, deep learning approaches, and orthodox stacking ensembles.
Ablation studies confirm these gains stem from structured feature decomposition
rather than increased complexity, particularly when groups contain complementary,
weakly redundant information (as supported by our conditional mutual information
analysis). Furthermore, STRIKE's modularity improves transparency, as groupwise
performance directly reveals how different feature sources contribute to risk
prediction, facilitating analysis in regulated financial environments.

% Future work may explore automated strategies for discovering optimal feature
% group structures directly from data, as well as extending the meta-learning
% stage to selectively model sparse cross-group interactions. Another promising
% direction is applying STRIKE to other heterogeneous tabular domains such as
% fraud detection, healthcare risk modeling, and recommender systems, where
% multiple feature sources often provide complementary predictive signals.

Future work includes automating feature group discovery, extending the
meta-learning stage to capture sparse cross-group interactions, and applying
STRIKE to other heterogeneous tabular domains like fraud detection and
healthcare risk modeling.

\begin{credits}
\subsubsection{\ackname} 
The authors acknowledge the University of Maryland supercomputing resources (http://hpcc.umd.edu), specifically the Zaratan cluster, made available for conducting the research reported in this paper.

\subsubsection{\discintname}
The authors have no competing interests to declare that are relevant to the content of this article.
\end{credits}

\bibliographystyle{splncs04}

\bibliography{references}

\end{document}